\documentclass[11pt,a4paper]{article}
\usepackage{CJKutf8}
\pdfoutput=1
\usepackage[hyperref]{rocling2021}
\usepackage{times}
\usepackage{latexsym}
\usepackage{amsmath}
\usepackage{graphicx}

\usepackage{tikz}
\usepackage{pgfplots}
\pgfplotsset{width=7.5cm,compat=1.12}
\usepgfplotslibrary{fillbetween}

\usepackage{microtype}

\roclingfinalcopy 
\def\roclingpaperid{014} 

\title{Integrated Semantic and Phonetic Post-correction \\for Chinese Speech Recognition}

\author{Yi-Chang Chen \\
  {\small E.SUN Financial Holding Co., Ltd.} \\
  \texttt{\small ycc.tw.email@gmail.com} \\\And
  Chun-Yen Cheng \\
  {\small E.SUN Financial Holding Co., Ltd.} \\
  \texttt{\small quadratic999@gmail.com} \\\And
  Chien-An Chen \\
  {\small E.SUN Financial Holding Co., Ltd.} \\
  \texttt{\small lukechen419@gmail.com} \\\AND
  Ming-Chieh Sung \\
  {\small E.SUN Financial Holding Co., Ltd.} \\
  \texttt{\small mingchieh-17908@email.esunbank.com.tw} \\\And
  Yi-Ren Yeh \\
  {\small Department of Mathematics} \\
  {\small National Kaohsiung Normal University}\\
  \texttt{\small yryeh@nknu.edu.tw} \\
}

\date{}

\begin{document}
\begin{CJK*}{UTF8}{bsmi}
\maketitle

\begin{abstract}
Due to the recent advances of natural language processing, several works have applied the pre-trained masked language model (MLM) of BERT to the post-correction of speech recognition. However, existing pre-trained models only consider the semantic correction while the phonetic features of words is neglected. The semantic-only post-correction will consequently decrease the performance since homophonic errors are fairly common in Chinese ASR. In this paper, we proposed a novel approach to collectively exploit the contextualized representation and the phonetic information between the error and its replacing candidates to alleviate the error rate of Chinese ASR.\footnote{The code, dataset and pre-trained models are available at \url{https://github.com/GitYCC/phonetic_mlm}} Our experiment results on real world speech recognition datasets showed that our proposed method has evidently lower CER than the baseline model, which utilized a pre-trained BERT MLM as the corrector.
\end{abstract}


\section{Introduction}


A variety of real-world applications have been benefited from the recent advances of automatic speech recognition (ASR), such as voice-activated banking, meeting minutes transcription, and voice content inspection. In ASR, hidden Markov model (HMM) based models \citep{Rabiner:86,Rabiner:89,Povey:11} and end-to-end models \citep{LAS,Bahdanau:16,RNN-T,Jaitly:16} are two popular types of modeling methods. For end-to-end models, it typically requires a huge amount of data for the model training due to the complicated architectures of neural networks. However, it is not easy to collect sufficient voice data in many real-world scenarios.


In contrast to end-to-end models, conventional HMM-based models, such as Kaldi \citep{Povey:11}, require less data and are quite popular in practice. HMM-based models are comprised of the acoustic model and language model. The acoustic model is used to produce phonetic units from the speech signals. Language models are responsible for obtaining the probabilities of next words by given past words. Typically the N-gram model is used as the language model in HMM-based models. One drawback of the N-gram model is the lack of long-term contextual clues by comparing with RNN-based or transformer-based language models. 

For Chinese speech recognition, we found that many homo-phonic errors are produced in HMM-based models with the N-gram model. It shows that the n{\"a}ive N-gram model might sacrifice the performance of HMM-based models even a well-trained acoustic model is given. However, it is not easy to replace the N-gram model due to the structure of interaction between the acoustic model and language model within HMM-based models. To overcome this problem, many methods have been proposed for the post-correction of speech recognition \citep{Shankar,Ziang, Jinxi,Xiaodong,Shaohua}. 


Recently, many successful methods have been proposed in natural language processing, such as BERT \citep{BERT}. For those pretraining tasks in BERT, masked language modeling (MLM) is a task of interest for our post-correction. The goal of MLM is to predict those masked tokens within a sentence in which certain input tokens are randomly masked. The prediction of masked tokens can be regarded as a kind of error correction. As shown in \citep{BERT}, MLM also could be applied as a post-correction for speech recognition. To be more precise, we apply the fine-tuned BERT to detect the errors within a recognized sentence from ASR. Followed by the detection, MLM is applied to correct these words.


The post-correction by MLM could reduce the deficiency of long-term contextual information in the N-gram model. However, the conventional MLM did not take the phoneme into account. To address this issue, we aim to propose a phonetic MLM as the post-correction for speech recognition by leveraging the phoneme information from the predicted words.



\section{Related Work}
Many methods have been proposed for correcting the outputs of ASR systems \citep{overview}. These existing approaches of language correction typically can be divided into three categories. The first group of them uses external language models to rescore k-best candidates in ASR system. For example, \citep{Shankar} picks k-best candidates of each word from the original ASR system. Once these k-best candidates are determined, RNN-LM is applied to re-score the k-best candidates of each word. From \citep{Shankar}, it also shows that the improved performance can be achieved since RNN-LM is a more effective model for the representation of natural languages.



The second category of language correction methods adopts the sequence to sequence learning framework \citep{S2S}. Based on this architecture, \citep{Ziang} adopts a character-based attention mechanism to generate a corrected sentence. On the other hand, \citep{Jinxi} also proposes a RNN with attention to correct the output from Listen, Attend, and Spell (LAS) model. 

The third group of language correction methods adopts a two-step correction. For example, \citep{Xiaodong} uses the language model and statistical machine translation model to detect error words in a sentence. After the error detection, SVM is used to replace the predicted error words with the most likely word. In \citep{Shaohua}, the authors proposed a bi-GRU model as the error detection network. Given a sequence of embeddings from BERT, the detection networks generate the probability of being an incorrect word. Followed by the detection network, the input of the correction model is the convex combination of mask token embedding and token embedding with the probability of incorrectness. Once the integrated embedding is calculated, a sequential multi-class labeling model based on BERT is applied to generate the corrected sentence. 



\section{Methodology}

In our proposed method, we integrate semantic and phonetic information for the post-correction of ASR. More specifically, the mask language model (MLM) based on BERT is used for semantic error correction. Besides, we also apply a phonetic distance to re-rank the candidates of being corrected from MLM. The details will be addressed in Section \ref{mask_prediction} and Section \ref{p_mask_prediction} respectively.


\subsection{Semantic Post-correction by MLM}
\label{mask_prediction}
\begin{figure*}[t]
\centerline{\includegraphics[width=13cm]{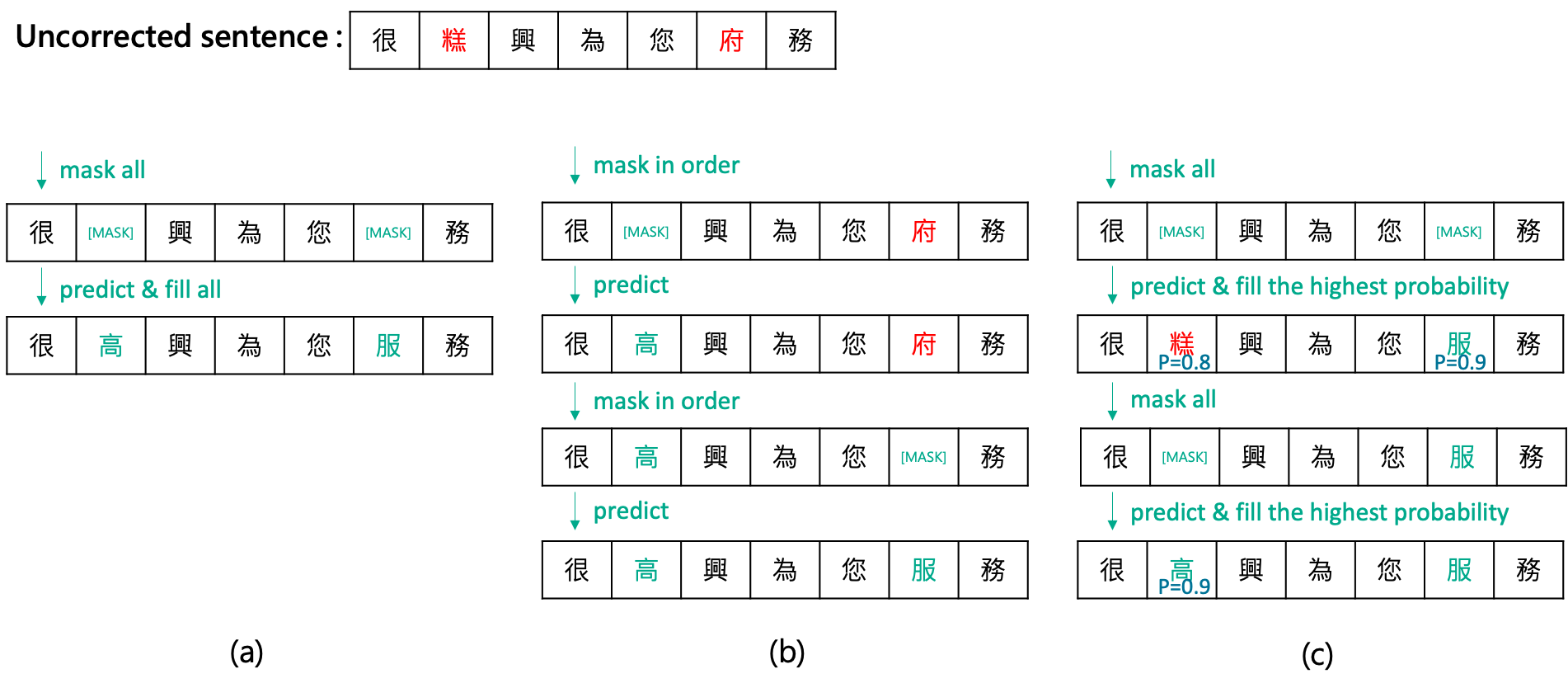}}
\caption{Different masking and replacement strategies of MLM for post-correction: (a) mask-all-and-replace-all, (b) mask-one-and-replace-one, and (c) mask-all-and-replace-one.}
\label{fig_four_MLM}
\end{figure*}

In our work, we first apply a token classifier to detect the errors within a recognized sentence from ASR. To learn the binary classifier, we regard the incorrect words within a sentence as the positive examples and fine-tune the model with Chinese pre-trained BERT. Followed by the detection, MLM is applied to correct these words. MLM is one of the pre-training tasks of BERT and originally aims to predict those masked tokens within a sentence in which certain input tokens are randomly masked. In the original design for the pre-training BERT, MLM predicts all masked tokens (i.e., the error words in our task) in a sentence simultaneously as shown in Figure~\ref{fig_four_MLM}(a). That is, the mask-all-and-replace-all strategy applied the error token classifier to detect all candidates of incorrect words. Once the detected error words are determined by the token classifier, we replace all of them by the ``[MASK]" token and predict the correct words by MLM at the same time. 

In addition to the mask-all-and-replace-all strategy, we also propose two other strategies to investigate the influence of the sequential masking and replacement of the detected error words. Different to mask-all-and-replace-all, our first strategy, mask-one-and-replace-one as shown in Figure~\ref{fig_four_MLM}(b), applies MLM to predict the correct words for each error token sequentially from left to right after the positions of error tokens are determined.


Similar to mask-all-and-replace-all, our second strategy, mask-all-and-replace-one, also masks all the candidates at the beginning. Rather than replace all the candidates at once, only one candidate associated with the highest probability will be replaced at one time as shown in Figure~\ref{fig_four_MLM}(c).

    
    


Based on the strategies mentioned above, the edited sentence will go through the same process all over again until all detected error words has been corrected. In our experiments, we also evaluate the performance of using these different strategies. The detailed results will be discussed in Section \ref{exp_mask_prediction}.

\subsection{Phonetic MLM for Post-correction}
\label{p_mask_prediction}

\begin{figure*}[t]
\centerline{\includegraphics[height=7cm]{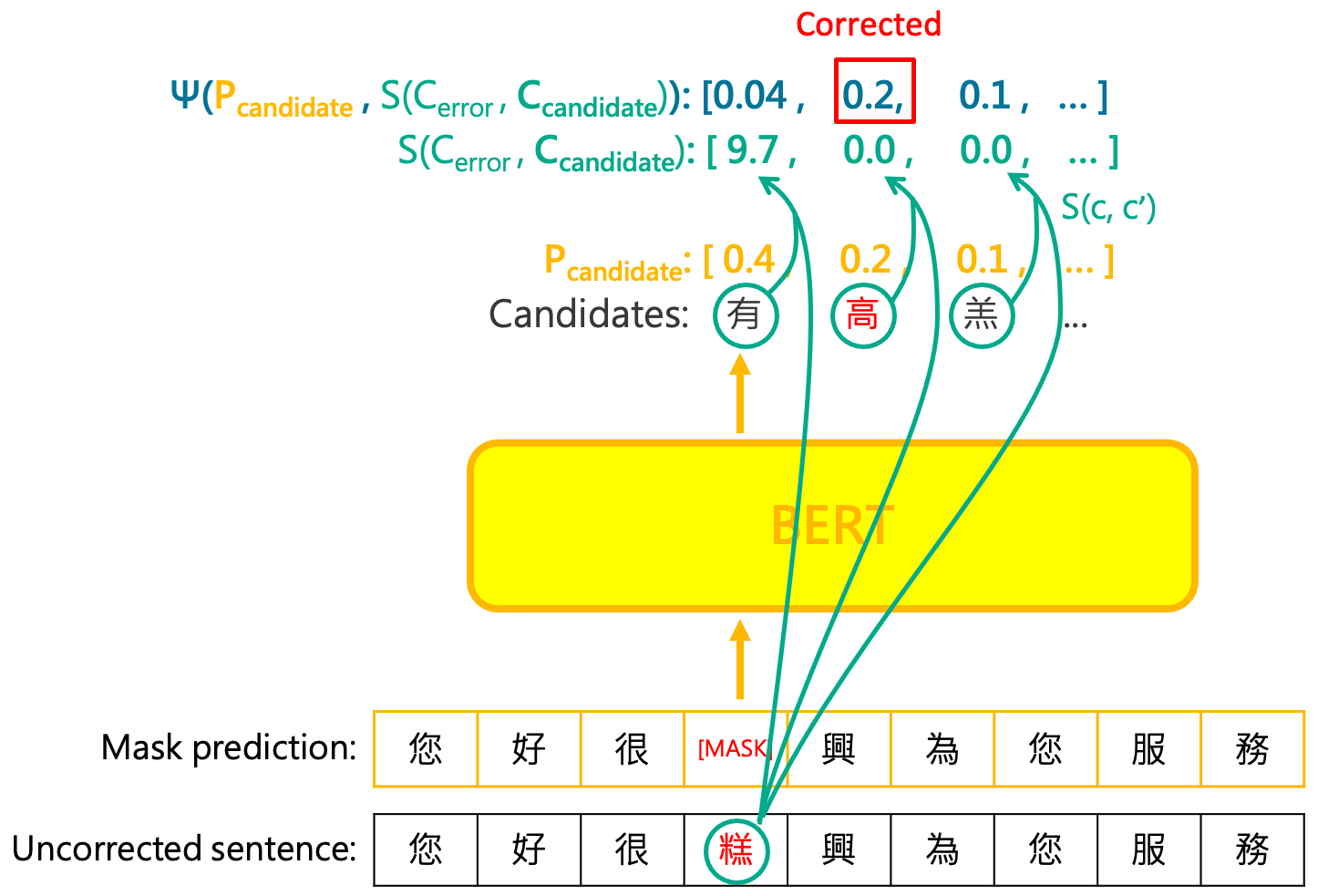}}
\caption{An example of our proposed semantic and phonetic post-correction. $P_{candidate}$ is the probabilities of candidates from MLM. $S(c_{error}, c_{candidate})$ is the the phonetic distances between the detected error character of interest ($c_{error}$) and the candidates ($c_{candidate}$) based on (\ref{p_dis}). $\Psi(\cdot,\cdot)$ controls the trade-off between semantic and phonetic metrics as defined in (\ref{eq_proposed_func_1}).}
\label{fig1}
\end{figure*}


Using conventional MLM as post-correction of speech recognition only takes the semantic context into account. As the example recognized sentences shown in Figure \ref{fig1}, we found that many homo-phonic errors of correction are made in HMM-based models with the N-gram language model. To overcome this problem, we proposed a phonetic MLM by leveraging the phonetic distance to integrate semantic and phonetic information for the post-correction.


In our proposed framework as shown in Figure~\ref{fig1}, we first apply the fine-tuned BERT of token classification to detect the positions of errors. Once the errors are determined, we simply mask them and apply MLM to get the probabilities of candidates denoted by $P_{candidate}$. As the example in Figure~\ref{fig1}, we first detect the error ``糕" in the recognized sentence, and then ``糕" is replaced by ``[MASK]". After masking ``糕", our MLM will predict candidates of replacement, such as ``有", ``高", and ``羔", with the corresponding probabilities $0.4$, $0.2$, and $0.1$ respectively.


In addition to the semantic correction by the conventional MLM, we also take the phonetic information into account. To obtain the phonetic information, we apply DIMSIM \citep{li-etal-2018-dimsim} to obtain the Chinese phonetic distance. In DIMSIM, each pronunciation of Chinese characters is encoded in a high dimensional space. The phonetic distance $S$ between Chinese characters $c$ and $c'$ is defined as follows:
\begin{equation}
\label{p_dis}
\begin{split}
& S(c,c')= \\
& S_{p}(p_{c}^I,p_{c'}^I)+S_{p}(p_{c}^F,p_{c'}^F)+S_{T}(p_{c}^T,p_{c'}^T),
\end{split}
\end{equation}
where $p^I_c$, $p^F_c$ and $p^T_c$ represent the initial, final, and tone components of $c$ in Pinyin, respectively. $S_p$ and $S_T$ are denoted as the Euclidean distance and phonetic tone distance between $c$ and $c'$, respectively. We note that the phonetic distance $S$ between two homo-phonic characters is $0$, and the phonetic distance $S(c,c')\geq 0$. In (\ref{p_dis}), by given two Chinese characters, the phonetic distance will be larger while the phonic difference is more significant.



Based on (\ref{p_dis}), we could calculate the phonetic distances between the detected error character of interest ($c_{error}$) and the candidates ($c_{candidate}$) of replacing $c_{error}$ by $S(c_{error}, c_{candidate})$. For example, we will calculate $S(\mbox{``糕"}, \mbox{``有"})$, $S(\mbox{``糕"}, \mbox{``高"})$, and $S(\mbox{``糕"}, \mbox{``羔"})$ as their phonetic distances in Figure \ref{fig1}. To consider the semantic correction and phonetic distance for the selection of candidates simultaneously, we first estimate $P_{candidate}$ of all candidates by MLM. Once $P_{candidate}$ and $S(c_{error}, c_{candidate})$ are obtained, we balance these two metrics by the function $\Psi$ as follows:
\begin{equation}
\begin{split}
& \Psi(P_{candidate}, S(c_{error}, c_{candidate})) \\
& =P_{candidate}\times exp(-\alpha\times S(c_{error}, c_{candidate})),
\end{split}
\label{eq_proposed_func_1}
\end{equation}
where $\alpha$ is a positive number that controls the trade-off between semantic and phonetic information. In our experiments, this hyperparameter is determined by grid search with a validation set. As the example in Figure~\ref{fig1}, given the error of interest (i,e., ``糕"), $S(\mbox{``糕"}, \mbox{``有"})$, $S(\mbox{``糕"}, \mbox{``高"})$, and $S(\mbox{``糕"}, \mbox{``羔"})$ are calculated as $9.7$, $0.0$, and $0.0$ by (\ref{p_dis}), respectively. For the correction, we use (\ref{eq_proposed_func_1}) to obtain the final scores $0.04$, $0.2$, and $0.1$ for ``有", ``高", and ``羔", respectively. Based on the scores from (\ref{eq_proposed_func_1}), we chose the character with the highest score as the replacement (i.e., ``高" in Figure \ref{fig1}).


\begin{table*}[t]
\begin{center}
\begin{tabular}{lrr}
\hline
\multicolumn{1}{c}{~} & \multicolumn{2}{c}{Datasets} \\
\multicolumn{1}{c}{~} & \multicolumn{1}{c}{AISHELL-3} & \multicolumn{1}{c}{Wiki} \\
\hline
mask-all-and-replace-all & 11.69 \% & 75.14 \% \\
mask-one-and-replace-one & 9.89 \% & 73.84 \% \\
mask-all-and-replace-one & 11.75 \% & 75.62 \% \\
\hline
\end{tabular}
\caption{The correction accuracies for different masking and replacement strategies.}
\label{table-mask_prediction-result}
\end{center}
\end{table*}

\begin{table*}[t]
\begin{center}
\begin{tabular}{l|rrr|rrr}
\hline
\multicolumn{1}{c|}{~} & \multicolumn{3}{c|}{Correction} & \multicolumn{1}{c}{} \\
\multicolumn{1}{c|}{~} & \multicolumn{1}{c}{Pre.} & \multicolumn{1}{c}{Rec.} & \multicolumn{1}{c|}{$F_1$} & \multicolumn{1}{c}{CER} \\
\hline
MLM & 0.099 & 0.061 & 0.075 & \multicolumn{1}{c}{10\%} \\
Ours ($\alpha=500$) & \textbf{0.404} & \textbf{0.179} & \textbf{0.248} & \multicolumn{1}{c}{\textbf{8.3\%}} \\
\hline
\end{tabular}
\caption{The evaluation results of our proposed method and the baseline model on AISHELL-3 dataset. Pre., Rec., $F_1$ represent the correction precision, recall and $F_1$-score denoted in \citep{tseng2015introduction}, respectively.}
\label{table-all-result}
\end{center}
\end{table*}

\section{Experiments}


Different to the conventional typo correction, we aim to correct the error after ASR in this work. To obtain the results of ASR, we use Kaldi \citep{Povey:11} as the speech recognizer in our experiments. Once the ASR results are generated, the correction methods are applied to refine the sentences. To evaluate our proposed methods, we conduct two experiments in this section. For the first one, we evaluate the performance of the semantic-only post-correction with MLM in Section \ref{mask_prediction}. In the second experiment, our proposed semantic and phonetic post-correction in Section \ref{p_mask_prediction} is also evaluated. The details will be addressed in the following sections.



\subsection{Evaluation on Semantic-only Post-correction}
\label{exp_mask_prediction}

In this experiment, we aim to evaluate the effectiveness on the semantic-only post-correction with MLM by considering different masking and replacement strategies as described in Section \ref{mask_prediction}. For the error detection, we assume that our detection network could detect all the incorrect words perfectly. Based on the setting, we calculate the accuracy of correction by given the detected incorrect characters. In our evaluation, we use two benchmark datasets in this experiment. The first one is a Chinese open speech dataset: AISHELL-3 \citep{aishell-3}. AISHELL-3 contains 63,262 and 24,773 sentences as the training set and test set respectively. It is worth noting that we directly use the pre-trained MLM of BERT with different masking strategies. Thus, we did not use the training set and only sampling 20,000 sentences from the testing set for the evaluation. The second one is Wiki dataset. The dataset contains 286,975 sentences, and all of them are used for the evaluation.

From the evaluation on Wiki dataset, as the results are shown in Table \ref{table-mask_prediction-result}, the mask-one-and-replace-one strategy produces the lowest accuracy. This indicates that if we only mask one incorrect character, the other unmasked incorrect characters will sacrifice the performance of MLM. On the other hand, if the incorrect characters are all masked, such as mask-all-and-replace-all and mask-all-and-replace-one strategies, the incorrect semantic information will not propagate to the task of token replacement. For AISHELL-3 dataset, we also can obtain similar results from the evaluation even if there are a lot of proper nouns in the sentences. Besides, the results from Table \ref{table-mask_prediction-result} also show that mask-all-and-replace-all and mask-all-and-replace-one strategies produce similar results for the token correction. For the sake of simplicity, we applied the mask-all-and-replace-all strategy in our experiment as the origin MLM of BERT did.



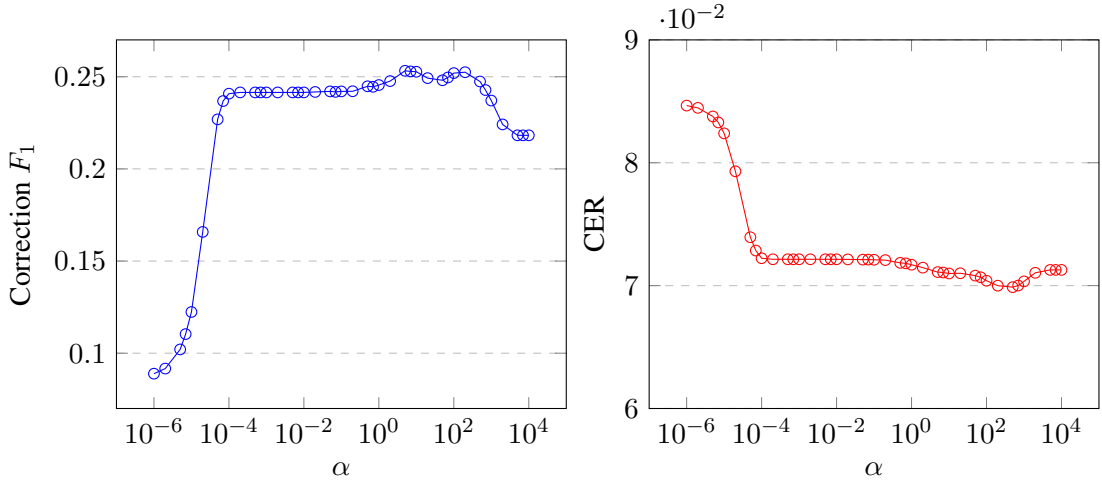
\begin{figure*}[t]
\begin{tikzpicture}
\begin{axis}[
    title=,
    xmode=log,
    xlabel={$\alpha$},
    ylabel={Correction $F_1$},
    ymin=0.07, ymax=0.27,
    xtick={0.000001, 0.0001, 0.01,1.0,100,10000},
    ytick={0.1,0.15,0.2,0.25,0.30},
    legend pos=south east,
    ymajorgrids=true,
    grid style=dashed,
]

\addplot[
    color=blue,
    mark=o,
    ]
    coordinates {
(1.00E-06, 0.088888889)
(2.00E-06, 0.091658342)
(5.00E-06, 0.102051026)
(7.00E-06, 0.110374171)
(1.00E-05, 0.122367101)
(2.00E-05, 0.165804416)
(5.00E-05, 0.2268788)
(7.00E-05, 0.236818003)
(0.0001, 0.240821161)
(0.0002, 0.241497495)
(0.0005, 0.241497495)
(0.0007, 0.241497495)
(0.001, 0.241497495)
(0.002, 0.241497495)
(0.005, 0.241497495)
(0.007, 0.241497495)
(0.01, 0.241497495)
(0.02, 0.241761139)
(0.05, 0.242056691)
(0.07, 0.241793013)
(0.1, 0.242056691)
(0.2, 0.242120533)
(0.5, 0.244822583)
(0.7, 0.244553017)
(1, 0.245502646)
(2, 0.247646825)
(5, 0.253232041)
(7, 0.252907365)
(10, 0.252713386)
(20, 0.249259758)
(50, 0.248101952)
(70, 0.249626511)
(100, 0.251977093)
(200, 0.252477974)
(500, 0.247419805)
(700, 0.242781049)
(1000, 0.237082923)
(2000, 0.224137931)
(5000, 0.218228135)
(7000, 0.218228135)
(10000, 0.218228135)
    };
    
\end{axis}
\end{tikzpicture}
\begin{tikzpicture}
\begin{axis}[
    title=,
    xmode=log,
    xlabel={$\alpha$},
    ylabel={CER},
    ymin=0.06, ymax=0.09,
    xtick={0.000001,0.0001,0.01,1.0,100,10000},
    ytick={0.06,0.07,0.08,0.09},
    legend pos=south east,
    ymajorgrids=true,
    grid style=dashed,
]

\addplot[
    color=red,
    mark=o,
    ]
    coordinates {
(1.00E-06, 0.084653261)
(2.00E-06, 0.084467909)
(5.00E-06, 0.083769833)
(7.00E-06, 0.083280475)
(1.00E-05, 0.082393302)
(2.00E-05, 0.079303853)
(5.00E-05, 0.073938892)
(7.00E-05, 0.072853557)
(0.0001, 0.072229809)
(0.0002, 0.07214426)
(0.0005, 0.07214426)
(0.0007, 0.07214426)
(0.001, 0.07214426)
(0.002, 0.07214426)
(0.005, 0.07214426)
(0.007, 0.07214426)
(0.01, 0.07214426)
(0.02, 0.072135026)
(0.05, 0.072116557)
(0.07, 0.07211747)
(0.1, 0.072104683)
(0.2, 0.07207698)
(0.5, 0.071853318)
(0.7, 0.071814197)
(1, 0.071702198)
(2, 0.071463187)
(5, 0.07109998)
(7, 0.071066486)
(10, 0.070987725)
(20, 0.07100647)
(50, 0.070814178)
(70, 0.070679176)
(100, 0.070399741)
(200, 0.069995211)
(500, 0.069875638)
(700, 0.070006042)
(1000, 0.070343839)
(2000, 0.071044176)
(5000, 0.071279948)
(7000, 0.071279948)
(10000, 0.071279948)
    };

\end{axis}
\end{tikzpicture}
\caption{Comparisons of correction $F_1$ and CER using different $\alpha$ in (\ref{eq_proposed_func_1}) for ALSHELL-3 dataset. }
\label{fig-grid-search}
\end{figure*}

\subsection{Evaluation on Our Semantic and Phonetic Post-correction}

In the second experiment, we evaluate our proposed phonetic MLM post-correction mentioned in Section \ref{p_mask_prediction} with only AISHELL-3 dataset since the phonetic information is not available in Wiki dataset. Different to the setting in Section \ref{exp_mask_prediction}, we randomly split 6,000 sentences from the training set as the validation set to find the proper hyper-parameters in our proposed method, and all the testing data are used for the evaluation. To evaluate the performance of the post-correction for ASR, we adopt correction $F_1$-score and CER (character error rate) as the metrics. Correction $F_1$-score is calculated by examining whether each error is corrected or not. Most Chinese error correction tasks adopt this metric as the evaluation \citep{tseng2015introduction}. On the other hand, CER is calculated by the average error rate in every sentence. It is often used to evaluate the results of speech recognition. To evaluate the performance in practices, we also report CER of the correction results in our experiments.


Followed by experimental results in Section \ref{exp_mask_prediction}, we use the pre-trained MLM model from the official bert-base-chinese package\footnote{\url{https://github.com/huggingface/transformers}} for the semantic correction. This semantic-only approach is also the baseline in this experiment. 
As shown in Table~\ref{table-all-result}, our proposed method could achieve 0.248 correction $F_1$-score while the baseline model only has 0.075 correction $F_1$-score. It shows that our proposed improve the performance of post-correction by leveraging the phonetic distance defined in (\ref{eq_proposed_func_1}). 

In addition to the correction $F_1$-score, we also evaluate the performance of these two models with CER due to the practical usage. Similar to the results with correction $F_1$-score, our proposed method also achieves better CER by comparing with the baseline model. Based on the results from Table~\ref{table-all-result}, we confirmed that the usage of phonetic information of characters is beneficial to post-correction of ASR. 




\subsection{Sensitivity of Phonetic Distance}
As discussed in Section \ref{p_mask_prediction}, we need to determine the hyper-parameter $\alpha$ in (\ref{eq_proposed_func_1}). This hyper-parameter controls the trade-off between semantic and phonetic information. In our experiments, we use the validation set to determine the value of $\alpha$ by the grid search. According to the range of phonetic distances from DIMSIM, we set $10^{-6}$ to $10^4$ as the search range, and calculate correction  $F_1$-score and CER with the validation data. Typically the larger $\alpha$ value we have, the more influence of the phonetic distance it will increase. As shown in Figure~\ref{fig-grid-search}, we plot the correction $F_1$-score and CER according to different values of $\alpha$. It can be observed that slightly increasing the value of $\alpha$ will improve the performance dramatically. This also indicates that many homo-phonic errors can be corrected by our proposed method. On the other hand, a too large value of $\alpha$ will also cause the opposite effect due to the over-emphasizing of phonetic information. Besides, it also shows that the results are quite robust within a wide range of $\alpha$. Thus, the proper value of $\alpha$ in (\ref{eq_proposed_func_1}) could be easily searched.

\subsection{Recoverable Ability of Phonetic Distance}
\begin{figure}[t]
\centerline{\includegraphics[width=7.5cm]{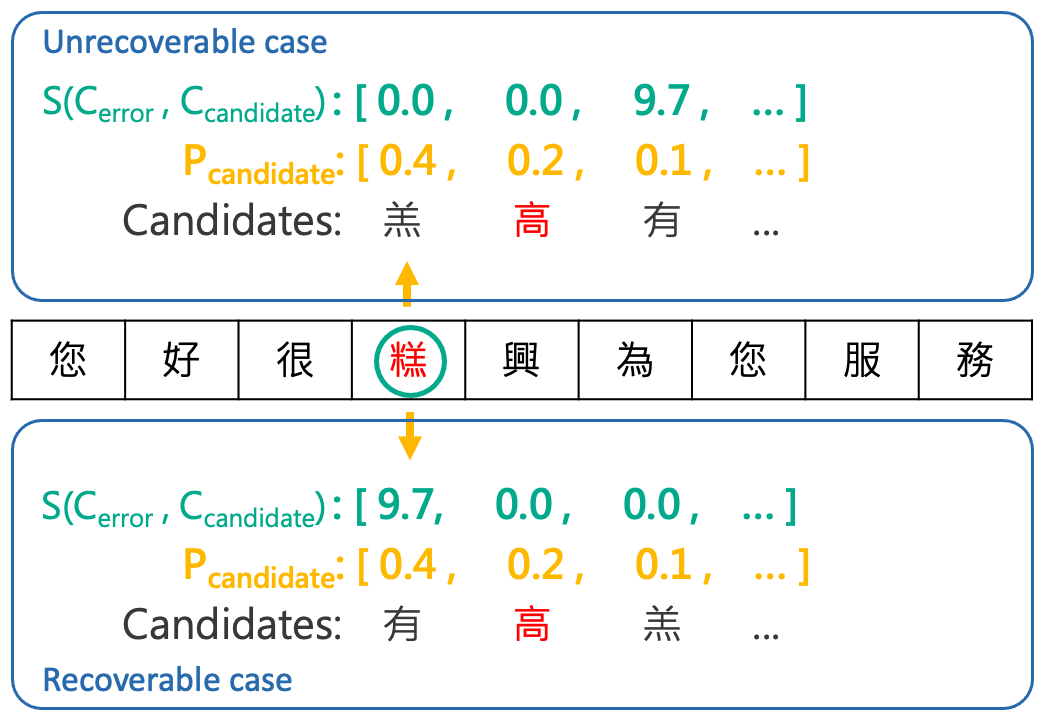}}
\caption{Examples of recoverable and unrecoverable cases in our scenario.}
\label{recoverable}
\end{figure}

In our proposed method, it is obvious that not all the incorrect characters can be corrected by adding the phonetic information. To be more precise, an error word of interest is unrecoverable if there exists a candidate that satisfies the following two conditions:
\begin{equation}
\begin{split}
P_{error~candidate}\geq P_{correct~candidate}
\end{split}
\label{unrecoverable_1}
\end{equation}
and
\begin{equation}
\begin{split}
& S(C_{error}, C_{error~candidate}) \\
\leq & S(C_{error}, C_{correct~candidate}),
\end{split}
\label{unrecoverable_2}
\end{equation}
where $C_{error}$ is the error word of interest, $C_{correct~candidate}$ is the ground truth, and $C_{error~candidate}$ is the incorrect word of the candidates. For example, as the unrecoverable case shown in Figure \ref{recoverable}, it is not possible to recover the correct character ``高" since ``羔" satisfies (\ref{unrecoverable_1}) and (\ref{unrecoverable_2}). On the other hand, one can recover the correct character ``高" as shown in the recoverable case of Figure \ref{recoverable} since no candidate satisfies (\ref{unrecoverable_1}) and (\ref{unrecoverable_2}). 

In our experiments, we have 21,865 Chinese characters that are not able to be corrected properly by the baseline model. Among these error corrections, we have 6,483 recoverable characters~($\sim$29.7\%). By given these recoverable characters, our proposed method can refine 4,671 characters ($\sim$72.1\%) correctly by using the phonetic distance. This indicates that our proposed phonetic feature could fix most recoverable characters.


\section{Conclusion}
In this paper, we proposed a novel approach for the post-correction of speech recognition. By exploring the phonetic distance derived from DIMSIM, we integrated semantic and phonetic information based on the pre-trained MLM of BERT. By taking the phonetic distance into account, many homophonic errors can be corrected by our proposed method. Experimental results on a real-world speech recognition dataset confirmed the use of our proposed method for improved post-correction of ASR. 

\section*{Acknowledgments}
This work was supported in part by the Ministry of Science and Technology of Taiwan under Grants MOST 108-2221-E-017-008-MY3.

\bibliography{rocling2021}
\bibliographystyle{acl_natbib}

\end{CJK*}
\end{document}